\documentclass[times,twocolumn,final]{elsarticle}

\usepackage{prletters}
\usepackage{lineno,hyperref}
\usepackage{subcaption}
\usepackage{multirow}
\usepackage{tabto}
\usepackage{amsmath,amssymb,amsfonts}
\usepackage{algorithmic}
\usepackage{graphicx}
\usepackage{textcomp}
\usepackage{xcolor}
\usepackage{multirow}
\usepackage{balance}
\usepackage{tikz}
\usepackage{color}
\usepackage{afterpage}
\usepackage{mathtools, cuted}
\usepackage[utf8]{inputenc}
\usepackage{caption}
\usepackage{upgreek}
\usepackage[ruled,vlined]{algorithm2e}
\usepackage{comment}

\DeclareMathAlphabet      {\mathbfit}{OML}{cmm}{b}{it}

\usepackage{soul}

\modulolinenumbers[5]

\journal{Pattern Recognition Letters}

\begin{document}

\begin{frontmatter}

\title{Large-Margin Representation Learning for Texture Classification}

\author[ets,uepg]{Jonathan de Matos\corref{mycorrespondingauthor}}
\cortext[mycorrespondingauthor]{Corresponding author}
\ead{jonathan@uepg.br}

\author[ufpr]{Luiz Eduardo Soares Oliveira}
\ead{luiz.oliveira@ufpr.br}

\author[pucpr,uepg]{Alceu de Souza Britto Junior}
\ead{alceu@ppgia.pucpr.br}

\author[ets]{Alessandro Lameiras Koerich}
\ead{alessandro.koerich@etsmtl.ca}

\address[ets]{\'Ecole de Technologie Sup\'erieure (\'ETS), Universit\'e du Qu\'ebec, 1100 Notre-Dame Street West, Montreal, QC, H3C 1K3, Canada}
\address[pucpr]{Pontificia Universidade Catolica do Paran\'a (PUCPR), Rua Imaculada Conceição, 1155, Curitiba, PR, 80215-901, Brazil}
\address[ufpr]{Universidade Federal do Paran\'a (UFPR), Rua Coronel Francisco Heraclito dos Santos, 100, Curitiba, PR, 81530-900, Brazil}
\address[uepg]{Universidade Estadual de Ponta Grossa (UEPG), Avenida General Carlos Cavalcanti, 4748, Ponta Grossa, PR, 84030-900, Brazil}

\begin{abstract}

This paper presents a novel approach combining convolutional layers (CLs) and large-margin metric learning for training supervised models on small datasets for texture classification. The core of such an approach is a loss function that computes the distances between instances of interest and support vectors. The objective is to update the weights of CLs iteratively to learn a representation with a large margin between classes. Each iteration results in a large-margin discriminant model represented by support vectors based on such a representation. The advantage of the proposed approach w.r.t. convolutional neural networks (CNNs) is two-fold. First, it allows representation learning with a small amount of data due to the reduced number of parameters compared to an equivalent CNN. Second, it has a low training cost since the backpropagation considers only support vectors. The experimental results on texture and histopathologic image datasets have shown that the proposed approach achieves competitive accuracy with lower computational cost and faster convergence when compared to equivalent CNNs. 

\end{abstract}

\end{frontmatter}


\section{Introduction}
\label{sec:introduction}
Convolutional neural networks (CNNs) have been established as the state-of-the-art approach to computer vision. CNNs achieve high accuracy due to how the convolutional layers filter images, the number of parameters available to learn representations \cite{resnet}\cite{inception}, and the large datasets \cite{imagenet} used in their training. They usually present high accuracy in object recognition problems, e.g., cars, people, animals, numbers. Networks pretrained with large datasets of objects can be reused in other contexts. There are two ways of transferring networks between contexts, fine-tuning or using them as feature extractors. The fine-tuning procedure allows fast training with fewer data because pre-trained filters are already able to identify some patterns. Although it reduces training effort, data may be insufficient even for fine-tuning in small datasets.
Furthermore, when working with datasets where textural information prevails to the detriment of shape and spatial characteristics, the first layers of pretrained CNNs may not respond well to new patterns. There are two additional problems: (i) deeper layers also have more problem-specific knowledge; (ii) there is the need for image size adaptation, which can slow the training~\cite{andrearczyk2016}. Using CNNs as feature extractors, there is no filter update. Instead, the activation maps of a specific layer are used as a feature vector for images of a new context, and an alternative classifier is trained on them. They work similarly to a handcrafted feature extractor and neither adapt to the new patterns nor generate features targeted to facilitate the classification task. Small and simple models like these are more suitable for classifying non-object and small datasets since they do not require data for representation learning but only for training a discriminant.

This paper proposes a novel large-margin representation learning that overcomes most of the problems mentioned above related to CNNs and handcrafted feature extractors. For this purpose, it addresses the following questions: a) is it possible to learn representations and discriminants on small-size texture datasets? b) is it possible to speed up the training convergence while achieving high accuracy? The proposed approach uses a short sequence of convolutional layers (CLs) to learn representation for texture classification. The CLs feed a large-margin discriminant that, in turn, provides information to update the CLs' weights and increase the decision margin. A novel loss function calculates the distance between instances in the decision frontier and anchors. The backpropagation algorithm minimizes the loss function while enlarging the margin between classes. The novelty of our approach is that it employs only instances that violate the decision margin to train the CLs and produce latent representations. That speeds up the convergence of the backpropagation algorithm to suitable latent representation and discriminant. In addition, it uses fewer parameters than a conventional CNN, allowing training in small datasets, and it performs well on non-object context recognition. Otherwise, the CNNs would still be more effective. The proposed approach was evaluated on a synthetic dataset based on Gaussian distributions, texture datasets, and three histopathological image (HI) datasets.

The main contributions of this paper are: (i) an approach that trains CLs from scratch with little data; (ii) A representation learning method that adapts itself to the characteristics of different texture datasets; (iii) a computational efficient training technique that uses only support vectors (SVs) in each iteration instead of all training instances; (iv) A fast convergence method compared to methods used for training conventional CNNs; (v) resilience to imbalanced data; (vi) A detailed comparison of the performance achieved by the proposed approach with approaches using handcrafted features and CNNs. 

This paper is organized as follows. Section~\ref{sec:background} presents some fundamental concepts. Section~\ref{sec:methods} describes the proposed approach and the experiment setup. In Section \ref{sec:results} , we present some experimental results and discuss the advantages of our method. The conclusions and the main contributions of our approach are presented in the last section.

\setlength{\belowdisplayskip}{0.3mm} \setlength{\belowdisplayshortskip}{0.3mm}

\vspace{-4mm}
\section{Related Works}
\label{sec:background}

The first CLs of CNNs are suitable to identify general and straightforward patterns such as textures. Their training help to make them adaptable to motifs of each context, so it is worthwhile to train these first CLs and use only them in more simple image contexts. One may find different contributions in the literature to adapt pre-trained CNN models to a new domain.

\citet{cimpoi2015} proposed the Fisher vector CNN (FV-CNN), which pools the last CL of a pretrained network, using it as a feature vector. The pooling allows using input images without resizing them to fit a fully connected CNN (FC-CNN). They used the FV-CNN features as input to an SVM and compared the FC-CNN and SIFT features. The FV-CNN showed to be a good texture descriptor, performing well on several benchmarks.
The texture CNN (TCNN)\citep{andrearczyk2016} uses a similar approach to the FV-CNN, but with the classification accomplished by a sequence of FC layers. Its difference to a traditional FC-CNN is the energy layer, a global average pooling (GAP) at the last CL whose outputs go to the FC layers. The SVM replacement by the FC layers as a classifier, in contrast to FV-CNN, permits training the CLs in conjunction with the classifier. This last approach does not require a pretrained CNN as it can train it with the FC layers, but it can explore transfer learning on low data scenarios, such as some types of medical images acquired by MRI, CT, radiography, or microscopy that have textural characteristics~\citep{heurtier2019}. Microscopy images allows the observation of tissues, their cells, and their nuclei, with a more texture oriented appearance. They are a challenging classification problem due to differences on staining process and small number of samples.
Therefore, instead of using large CNNs, one can use only their first layers that are more texture-aware as an FCN, intercepting the output and treating it as a feature vector. This latent representation can be used with a simpler classifier with fewer parameters than FC layers of a CNN. 

One can also use the deep metric learning (DML) concepts to train the FCNs to produce better filters and a representation more suitable to the new classifier.

Metric learning is an alternative to classification methods in situations where the number of classes is high, in the order of thousands, or the number of samples of each category is low \cite{chopra2005}. Two examples of this situation are face recognition and signature verification. The idea of the dissimilarity metric learning is to learn a representation to identify two samples' similarities, usually employing a distance metric, e.g., Euclidean distance. DML approaches allow metric and representation learning simultaneously. Learning representation is possible with deep learning due to the training of parameters used in its layers, mostly the convolutional ones. 
\citet{chopra2005} presented a method that uses siamese convolutional neural networks and the energy-based model. It consists of using raw energy values instead of probabilistic normalized ones. Their concept of energy is analogous to the one of \citet{andrearczyk2016}. The advantage of using CNNs is that they provide end-to-end training that learns low and high-level features and results in shift-invariant detectors. Their method aggregates the energy output of each siamese network into one neural network trained with contrastive loss. The loss allows training the system together (siamese networks and the single neural network), improving the sample representation on the energy layer for texture classification. 

DML algorithms can be split into pair-based or proxy-based. The former, like contrastive loss, aims to minimize intra-class distance and maximize inter-class distance. However, this approach has prohibitive computational complexity and pairs that do not contribute to the training. Therefore, several works have addressed these issues~\cite{proxyloss2020}. The ranked list loss method~\citep{wang2019} relies on a threshold margin that gives more attention, using weighting, to the samples that maximize the margins between opposite classes. The margin determines the negative points that are too close to the query and violates most of the margin. These are the negative selected points. On the other hand, proxy-based methods~\cite{movshovitz2017}~\cite{aziere2019} create an instance representing a set of instances of the same class. It reduces the number of training instances and avoids noise and outliers.

\vspace{-4mm}
\section{Large-Margin Fully Convolutional Network (LMFCN)}
\label{sec:methods}
The proposed approach is pair-based, and it selects the most effective instances for the training procedure, reducing the training complexity and discarding irrelevant instances. It also selects specific examples as anchors to calculate a novel loss function, which speeds up training. Although the proposed approach shares ideas of DML, our application context is different, with more instances per class and fewer classes than the usual DML context. 

The proposed method has three components: a fully convolutional network (FCN) with a global average pooling (GAP), a large-margin classifier, and a novel loss function, made up of three terms. The LMFCN\footnote{https://github.com/jonathandematos/lmfcn} uses a sequence of CL acting as a filter bank to learn representation from data used as input to a large-margin classifier. It acts as an end-to-end image classifier, like a CNN but requires less training data to learn high discriminant representations. Its advantage is to enable filter training and make the latent representation more suitable for the classifier. In addition, the backpropagation algorithm trains the filters with a particular loss function that uses the distance between instances of interest and their anchors provided by the large margin classifier to improve representation learning. 

\begin{figure}[ht!]
\centering
\begin{subfigure}{.15\textwidth}
  \centering
  \includegraphics[width=0.99\linewidth]{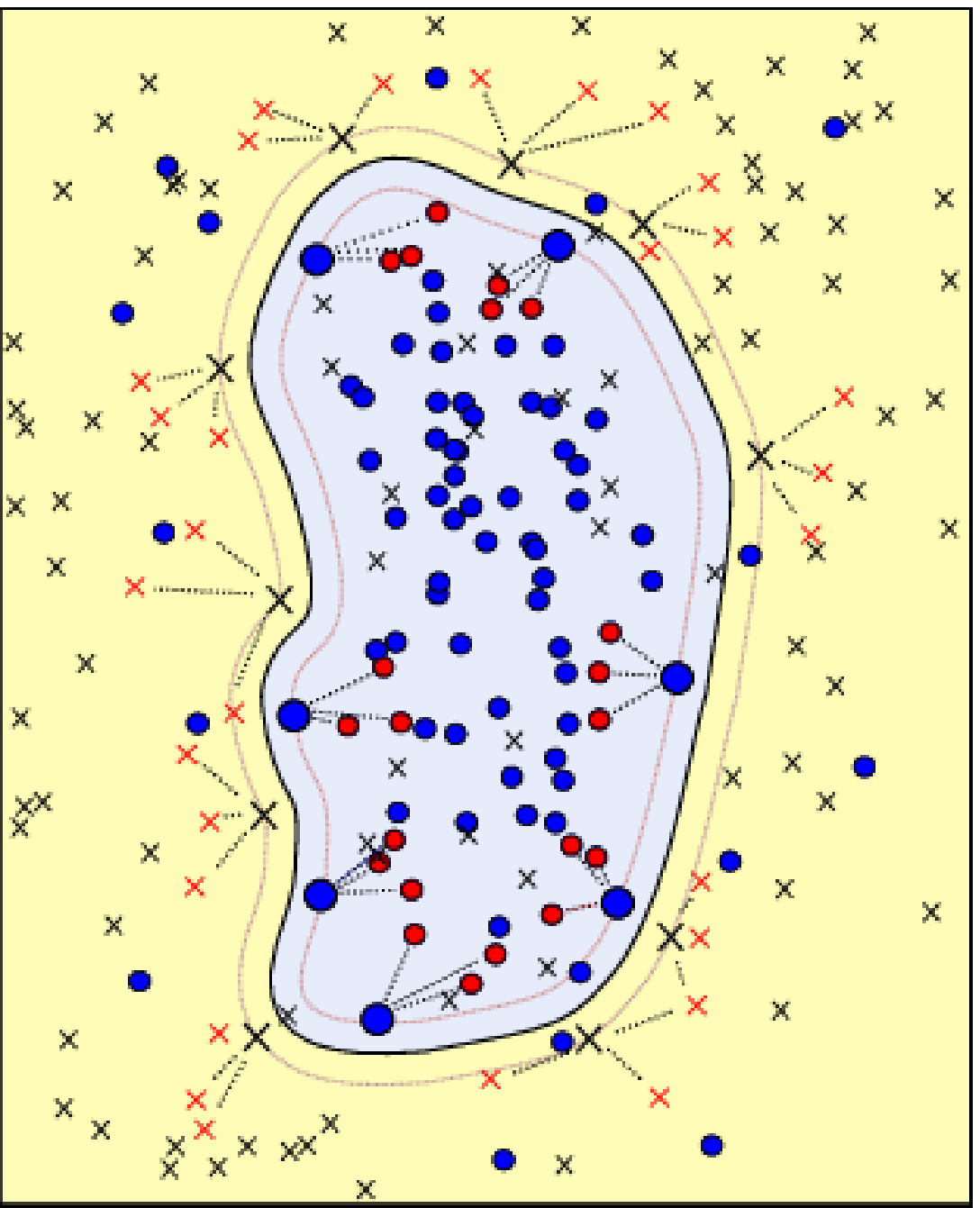}
  \vspace*{-5mm}
  \caption{Type 1 anchor}
  \label{fig:anchorstype1}
\end{subfigure}%
\begin{subfigure}{.15\textwidth}
  \centering
  \includegraphics[width=0.99\linewidth]{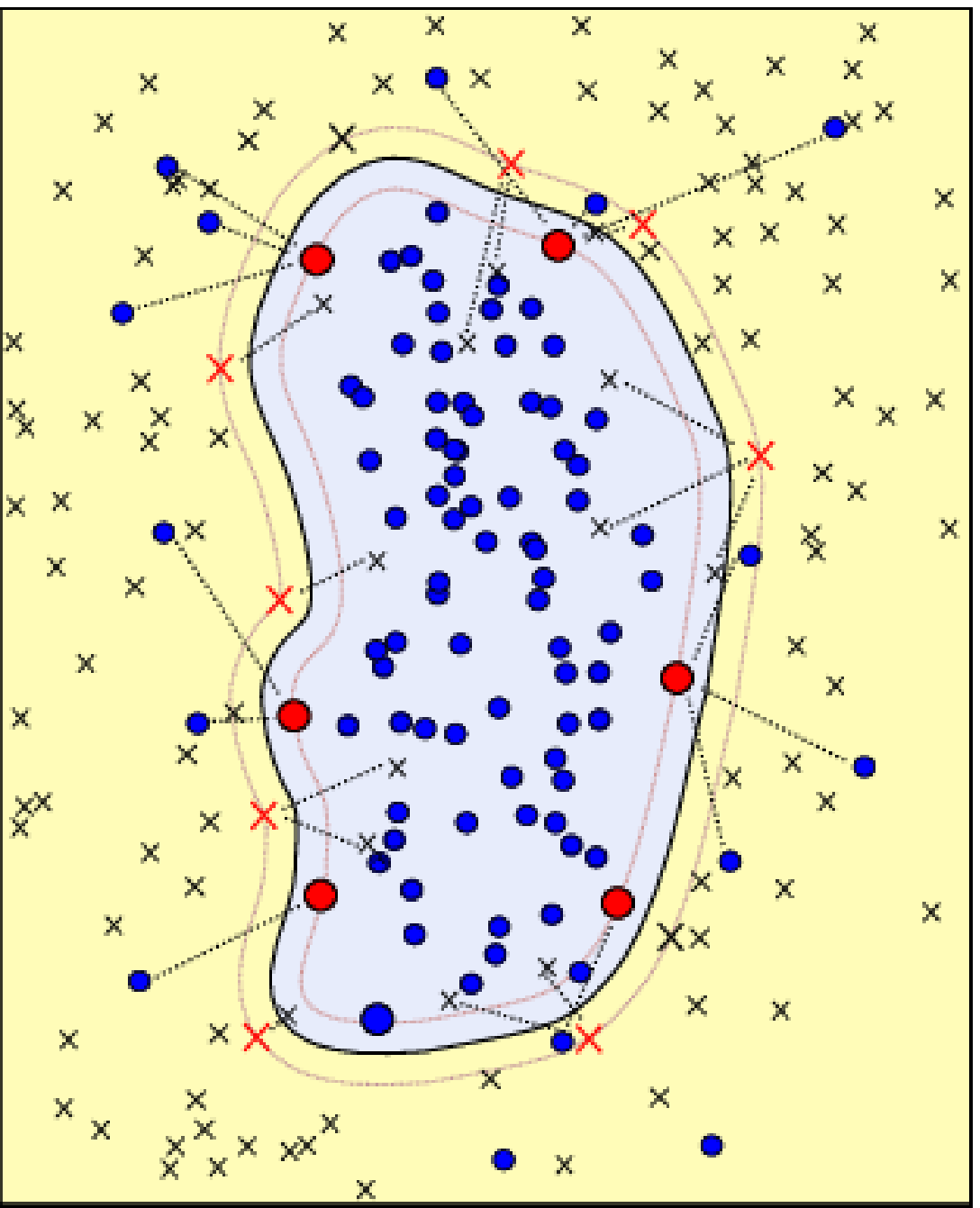}
  \vspace*{-5mm}
  \caption{Type 2 anchor}
  \label{fig:anchorstype2}
\end{subfigure}
\begin{subfigure}{.15\textwidth}
  \centering
  \includegraphics[width=0.99\linewidth]{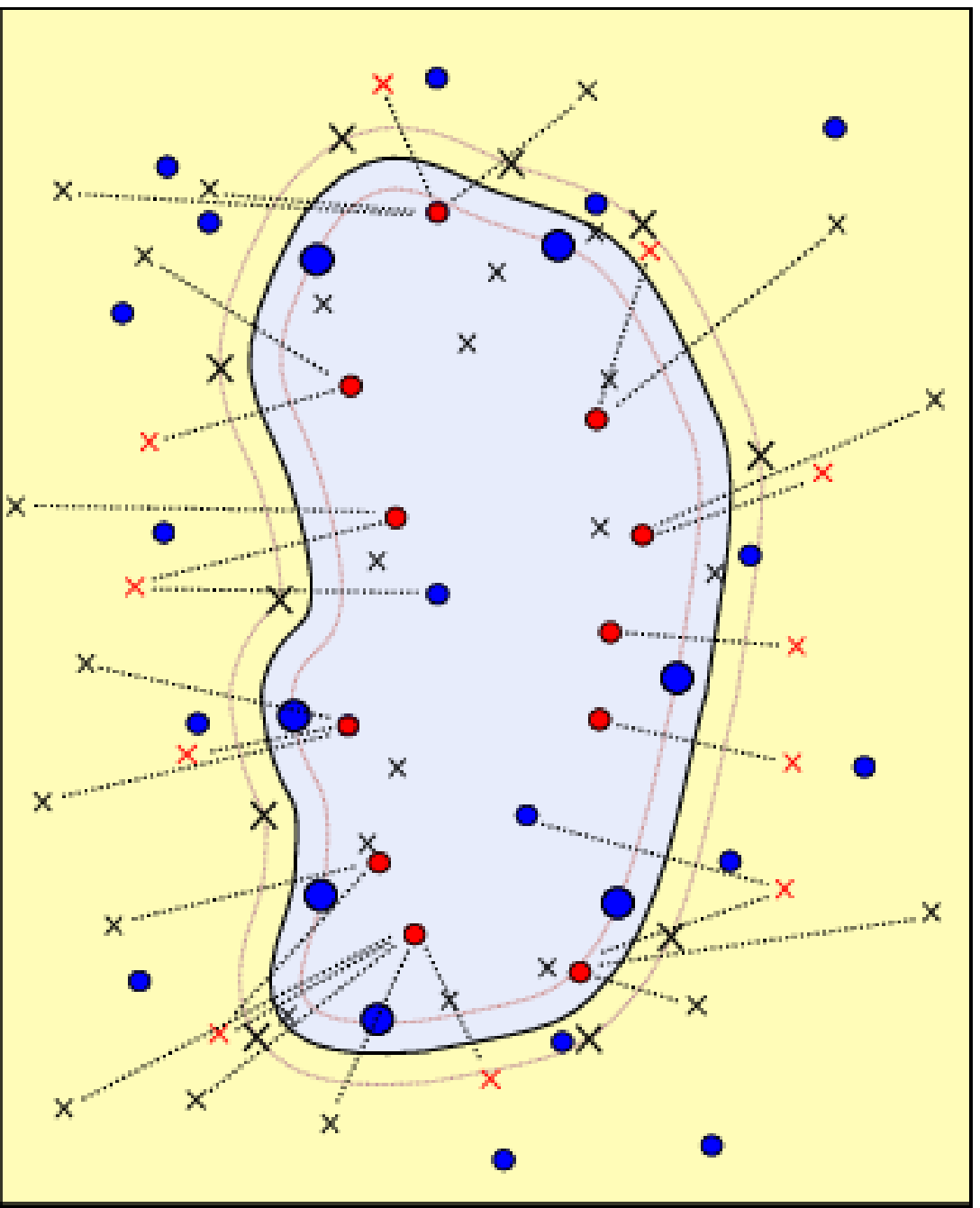}
  \vspace*{-5mm}
  \caption{Type 3 anchor}
  \label{fig:anchorstype3}
\end{subfigure}%
\vspace{-2mm}
\caption{A two-class latent space with decision boundary computed by an RBF SVM. Circles are samples of class 0 and crosses, class 1. Bigger symbols mean the SVs for each class. In blue is the region of class 0, and in yellow is class 1. Dashed straight black lines link the instances to their anchors. In red are the anchors for three different situations on the calculations. (a) Reference anchors to the SVs; (b) Anchors used to move the misclassified instances; (c) Anchors used to increase the separation of instances from opposite classes.}
\label{fig:anchors}
\vspace{-4mm}
\end{figure}

\subsection{Training Procedure and Loss Function} 

The learning algorithm uses the concept of anchors to guide the training. Anchors are instances from the training set used as references by the loss function to calculate the distance to instances of interest. The backpropagation algorithm minimizes such distances during training. Fig.~\ref{fig:anchors} presents a latent space split into two regions by an RBF SVM classifier and the three types of anchors.

Type 1 anchors (red circles and crosses) are the correctly classified instances closest to the support vectors (SVs). The LMFCN attempts to maximize the margin by pushing the SVs in the direction of such anchors. The learning algorithm minimizes the distance between them and the SVs. In Fig.~\ref{fig:anchorstype1}, the dotted straight lines linking SVs identify the three anchors of each SV and help visualize the effect of the distance minimization.

Type 2 anchors (Fig.~\ref{fig:anchorstype2}) are the SVs closest to misclassified examples used to move such instances in direction to the right side of the decision boundary.

Type 3 anchors are the closest correctly classified instances of the opposite class, as shown in Fig.~\ref{fig:anchorstype3}. Type 3 anchors help to maximize the distance between samples of different classes. The number of anchors, regardless of their type, is a hyperparameter.

\label{sec:algorithm}

\begin{figure}[htpb!]
    \centering
    \includegraphics[width=0.95\linewidth]{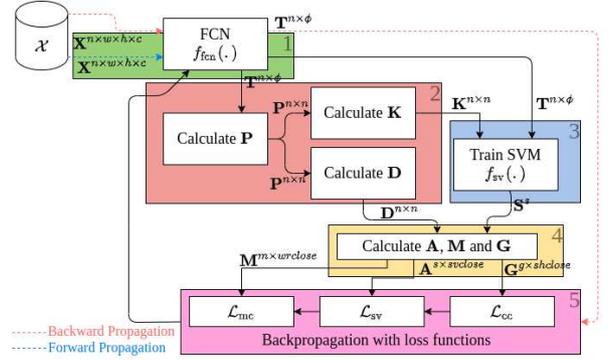}
    \vspace{-3mm}
    \caption{An overview of the training scheme of the LMFCN.}
    \label{fig:cnnsv}
    \vspace{-4mm}
\end{figure}

The training procedure starts with a dataset of size $\mathit{n}$ denoted as $\mathcal{X}=\{\mathbf{X}^{\mathit{t}}, \mathit{o^t}\}_{\mathit{t}=1}^{\mathit{n}}$, where $\mathit{t}$ indexes images $\mathbf{X}^{\mathit{t}}$ of width $\mathit{w}$, height $\mathit{h}$ and $\mathit{c}$ channels in $\mathcal{X}$, and $\mathit{o^t} \in \mathbb{N}:[0,1]$ is its expected output. All images from $\mathcal{X}$ are fed to CLs denoted as $\mathit{f_\text{fcn}(.)}$ and produces a matrix $\mathbf{T}^{\mathit{n}\times\phi}$, with $\phi$ being the size of the latent representation (step 1 at Fig.~\ref{fig:cnnsv}). The algorithm uses matrix $\mathbf{T}$ to calculate matrix $\mathbf{P}$ (Eq.~\eqref{eq:mtrp}), which in turn is used to calculate matrices $\mathbf{K}^{\mathit{n}\times\mathit{n}}$ and $\mathbf{D}^{\mathit{n}\times\mathit{n}}$ using Eqs.~\eqref{eq:kernel} and~\eqref{eq:distmatrix}, respectively (step 2 at Fig.~\ref{fig:cnnsv}). $\mathbf{K}$ is a RBF kernel matrix used to train a large margin classifier $\mathit{f_{sv}(.)}$, which produces the output $\mathit{y^t}$ for each element $\mathit{t}$ of $\mathcal{X}$, and also provides the set of support vectors (SVs) indexes $\mathbfit{S} = \{\mathit{s^u\}}_{\mathit{u=1}}^\mathit{v}$, where $\mathit{s^u} \subset \mathbb{N} : [0,\mathit{n}[$ and $\mathit{v}$ is the number of SVs (step 3 at Fig.~\ref{fig:cnnsv}). $\mathbf{D}$ is essential to the rest of the algorithm as it contains the pairwise distance of all instances and is used to create the anchor matrices.

\setlength{\abovedisplayskip}{-3pt}
\setlength{\belowdisplayskip}{6pt}
\setlength{\abovedisplayshortskip}{-3pt}
\setlength{\belowdisplayshortskip}{6pt}
\begin{align}
    \label{eq:mtrp}
    \mathit{p}_{ij} = \sum_{b=0}^{\phi}({T_{ib} - T_{jb}})^2
\end{align}
\begin{align}
    \label{eq:kernel}
    \mathit{k}_{ij} = \exp(-\gamma \mathit{p}_{ij}) 
\end{align}
\begin{align}
    \label{eq:distmatrix}
    \mathit{d}_{ij} = \sqrt{\mathit{p}_{ij}}
\end{align}

Type 1 anchors (step 4 at Fig.~\ref{fig:cnnsv}) are obtained from matrix $\mathbf{D}$, the set $\mathbfit{S}$, the expected output $\mathit{o^t}$, and the output from $\mathit{f_\text{sv}(f_\text{fcn}(}\mathbf{X}^\mathit{t}\mathit{))}$, as shown in Eq.~\eqref{eq:auxiliaryA}:

\begin{align}
\label{eq:auxiliaryA}
    \hspace{-5pt}\mathit{e}_{\mathit{ij}} = 
    \begin{cases}
        \tau&\text{if }\mathit{i=j}\textbf{ or }\mathit{j\in\mathbfit{S}}\text{ or }o^i\neq o^j\text{ or }o^j\neq f_{\text{sv}}(f_{\text{fcn}}(\mathbf{X}^j))\\
        \mathit{d}_{\mathit{ij}} & \text{otherwise}
    \end{cases}
\end{align}
\noindent where $\mathit{e}_{\mathit{ij}}$ is either the distance between SVs and their anchors ($d_{ij}$) or a large constant $\tau$ that indicates that there is no SV-anchor relation (uninteresting instances), $\mathit{f_\text{fcn}(}\mathbf{X}^\mathit{t}\mathit{)}$ denotes the latent representation generated by the FCN for an input $\mathbf{X}^\mathit{t}$, $\mathit{f_\text{sv}(.)}$ is the output predicted by the large-margin classifier.

Furthermore, we also define a sorting function $\mathit{f_\text{argsort}(.)}$ that takes a vector as input and returns a vector of indices sorted in increasing order. So, the anchor matrix $\mathbf{A}^{\mathit{s\times n}}$ is calculated using Eq.~\eqref{eq:type1matrix}.

\begin{align}
\label{eq:type1matrix}
\mathbf{a}_u = \mathit{f_\text{argsort}(}\mathbf{e}_{\mathit{s^u}}) \text{       with } \mathit{s^u} \in \mathbfit{S} \text{ and } \mathit{u} \in [0,|\mathit{S}|[
\end{align}
Likewise $\mathbf{A}$, we calculate with the Eqs.~\eqref{eq:type2matrix} and \eqref{eq:type3matrix} the matrices $\mathbf{M}^{\mathit{m \times n}}$ and $\mathbf{G}^{\mathit{g \times n}}$ for type 2 and 3 anchors, respectively (step 4 at Fig.~\ref{fig:cnnsv}). We define a set of indexes to the misclassified and correct classified instances from $\mathcal{X}$ as $\mathbfit{Q}=\{\mathit{q}\mid\mathit{q}\subset\mathbb{N}:[0,\mathit{n}[\wedge\mathit{f_\text{sv}(f_\text{fcn}(}\mathbf{X}^{\mathit{q}}\mathit{))}\neq\mathit{o^{q}}\}$, and $\mathbfit{R}=\{\mathit{r}\mid\mathit{r}\subset\mathbb{N}:[0,\mathit{n}[\wedge\mathit{r}\notin\mathbfit{S}\cup\mathbfit{Q}\}$, respectively.

\begin{align}
\label{eq:auxiliaryM}
    \mathit{z}_{\mathit{ij}} = 
    \begin{cases}
        \mathit{d}_{\mathit{ij}} & \text{ if } \mathit{i} \in \mathbfit{Q} \text{ and } \mathit{j} \in \mathbfit{S} \text{ and } \mathit{i} \neq \mathit{j}\\
        \tau & \text{ otherwise }
    \end{cases}
\end{align}
\begin{align}
\label{eq:type2matrix}
\mathbf{m}_i = \mathit{f_\text{argsort}(}\mathbf{z}_{\mathit{q^i}}) \text{ with } \mathit{q^i} \in \mathbfit{Q} \text{ and } \mathit{i} \in [0,|\mathbfit{Q}|[
\end{align}
\begin{align}
\label{eq:auxiliaryG}
    \mathit{h}_{\mathit{ij}} = 
    \begin{cases}
    \mathit{d}_{\mathit{ij}} & \text{ if } \mathit{i} \in \mathbfit{R} \text{ and } \mathit{j} \in \mathbfit{R} \text{ and } \mathit{o^i} \neq \mathit{o^j}\\
    \tau & \text{ otherwise }
    \end{cases}
\end{align}
\begin{align}
\label{eq:type3matrix}
\mathbf{g}_\mathit{i} = \mathit{f_\text{argsort}(}\mathbf{h}_{\mathit{r^i}}\mathit{)} \text{ with } \mathit{r^i} \in \mathbfit{R} \text{ and } \mathit{i} \in [0,|\mathbfit{R}|[
\end{align}

\noindent{where $\mathit{z}_{\mathit{ij}}$ and $\mathit{h}_{\mathit{ij}}$ in Eqs.~\eqref{eq:auxiliaryM} and \eqref{eq:auxiliaryG} are the distance between misclassified instances and the SVs ($d_{ij}$) and the distance between correct classified instances from opposite classes ($d_{ij}$), respectively. Again, $\tau$ is a large constant that indicates uninteresting instances.}

Step 5 of the algorithm is the backpropagation of images from $\mathcal{X^{\mathbf{S}}}$, $\mathcal{X^{\mathbf{Q}}}$ and $\mathcal{X^{\mathbf{R}}}$ through the FCN to calculate the gradients and use them with the loss functions to update the CL weights. Steps 1 to 5 define one epoch. After step 5, the process restarts from step 1, so the latent representation is computed again, as though the matrices {$\mathbf{P}$, $\mathbf{K}$ and $\mathbf{D}$} and the large-margin discriminant is retrained on such a updated latent representation, producing another set of SVs allowing recalculation of matrices $\mathbf{A}$, $\mathbf{M}$ and $\mathbf{G}$. Therefore, the elements of matrix $\mathbf{T}$ are different from the previous epoch because of the updated FCN weights. 

\label{sec:loss}

The proposed loss function relies on the similarity between examples, and it has three terms, as shown in Eq.~\eqref{eq:totalloss}.  It aims at finding a latent representation that maximizes the margin ($\mathcal{L}_\text{sv}$) while pushing misclassified examples towards the right side of the decision boundary ($\mathcal{L}_\text{mc}$) and moving well-classified examples farther away from the decision boundary ($\mathcal{L}_\text{cc}$).
\vspace{-2mm}

\begin{align}
    \mathcal{L}= \mathcal{L}_\text{sv} + \mathcal{L}_\text{mc} + \mathcal{L}_\text{cc}
    \label{eq:totalloss}
\end{align}
\vspace{-3mm}

$\mathcal{L}_\text{sv}$ calculates the sum of distances between SVs and its anchors using matrix $\mathbf{A}$,  
as shown in Eq.~\eqref{eq:l1}. As a consequence, 
the gradients are affected only by the instances which are SVs, not by the type 1 anchors, as they were already generated and stored in $\mathbf{T}$. The backpropagation procedure updates the weights in a way that the latent representation generated by $\mathit{f_\text{fcn}(.)}$ has the smallest possible distance to the fixed values of type 1 anchors at the current epoch. Therefore, the weights are updated to move the SVs and not the anchors.
\vspace{-3mm}

\begin{align}
    \mathcal{L}_\text{sv} = \frac{\displaystyle\sum_{i=0}^{|\mathcal{S}|} \sum_{j=1}^{sv_\text{close}} [\mathit{f_\text{fcn}}(\mathbf{X}^{\mathit{s^{i}}}) - \mathbf{t}_{a_{ij}}]^2}{|\mathbfit{S}|}
    \label{eq:l1}
\end{align}
\vspace{-3mm}

\noindent where $\mathit{sv_\text{close}}$ is the number of anchors to use for each SV, $\mathit{f_\text{fcn}(.)}$ is {the FCN updated by the backpropagation}, $\mathbf{X}^{\mathit{s^i}}$ is the matrix that represents the $\mathit{s^i}$ image from $\mathcal{X}$, $\mathit{a_{ij}}$ is an index pointing to an anchor instance in the input set $\mathcal{X}$, $\mathbf{t}_{a_{ij}}$ is the latent representation of an image $\mathbf{X}^{a_{ij}}$, {and $|\mathbfit{S}|$ is the number of SVs}.

The training algorithm computes the loss over the entire set of SVs as a single batch. It is also possible to use mini-batches, but considering small-size datasets and an FCN with compact architecture that yields a low-dimensional latent representation, this is unnecessary. In the next epoch, the updated weights will affect the generation of the latent representation of the entire dataset. Therefore, the latent representation of anchors also changes, and the training algorithm builds a new set of anchors.

$\mathcal{L}_\text{mc}$ calculates the summation of distances between misclassified instances and their anchors using matrix $\mathbf{M}$, as shown in Eq.~\eqref{eq:l2}. We also use all misclassified instances as a single batch, although there is no limitation to performing it in mini-batches. The influence of $\mathcal{L}_\text{mc}$ in the training algorithm is the weight updating that minimizes the distance between the misclassified instances and their closest SVs. Consequently, the misclassified instances are pushed towards the right side of the decision boundary.

\begin{align}
    \mathcal{L}_\text{mc} = \frac{\displaystyle\sum_{i=0}^{|\mathbfit{Q}|} \sum_{j=0}^{\mathit{wr_\text{close}}} [\mathit{f}_\text{fcn}(\mathbf{X}^{\mathit{q^i}}) - \mathbf{t}_{\mathit{m_{ij}}}]^2}{|\mathbfit{Q}|}
    \label{eq:l2}
\end{align}
\noindent where $\mathit{wr_\text{close}}$ is the number of anchors for each misclassified instance, $\mathit{f_\text{fcn}(.)}$ is {the FCN updated by the backpropagation}, $\mathbf{X}^{\mathit{q^i}}$ is the matrix that represents the $\mathit{q^i}$ image from $\mathcal{X}$, 
$\mathit{m^{ij}}$ is an index pointing to an anchor instance in the input set $\mathcal{X}$, $\mathbf{t}_{m_{ij}}$ is the latent representation of an image { $\mathbf{X}^{m_{ij}}$}, and {$|\mathbfit{Q}|$ is the number of misclassified instances}.

When looking at only a single misclassified instance, the weight updating may not be enough to move such an instance to the right side of the decision boundary because its anchors are SVs right at the margin of the decision boundary. Despite that, the misclassified example can become an SV in the following training epoch. 

$\mathcal{L}_\text{cc}$ represents the distance between well-classified instances and their anchors. Since we want to maximize such a distance, Eq.~\eqref{eq:l3} calculates the inverse of such a distance and incorporates it in the loss function, which is minimized during training.
\vspace{-3mm}

\begin{align}
    \mathcal{L}_\text{cc} = \frac{|\mathbfit{R}|}{\displaystyle\sum_{i=0}^{|\mathbfit{R}|} \sum_{j=0}^{\mathit{sh_\text{close}}} [\mathit{f}_\text{fcn}(\mathbf{X}^{\mathit{r^i}}) - \mathbf{t}_{\mathit{g_{ij}}}]^2}
    \label{eq:l3}
\end{align}
\vspace{-3mm}

\noindent where $\mathit{sh_\text{close}}$ is the number of anchors for each correctly classified instance,
$\mathit{f_\text{fcn}(.)}$ is {the FCN updated by the backpropagation}, $\mathbf{X}^{\mathit{r^i}}$ is the matrix that represents the $\mathit{r^i}$ image from $\mathcal{X}$, $\mathit{g^{ij}}$ is an index pointing to an anchor instance in the input set $\mathcal{X}$,
$\mathbf{t}_{g_{ij}}$ is the latent representation of an image $\mathbf{X}^{g_{ij}}$, and {$|\mathbfit{R}|$ is the number of correct classified instances}.


{The number of anchors used for each training instance is controlled by $\mathit{sv_\text{close}}$, $\mathit{wr_\text{close}}$, and $\mathit{sh_\text{close}}$ in Eqs.~\eqref{eq:l1}~\eqref{eq:l2}, and ~\eqref{eq:l3}, respectively. Usually, complex classification problems require a higher number of anchors. Furthermore, using all three terms of the proposed loss function in the training process may not always be necessary. $\mathcal{L}_\text{sv}$ alone already leads to good representations as such a loss is directly related to SVs and margin maximization. The computational cost for computing this term of the loss function is not high and decreases as the number of instances used by the backpropagation algorithm reduces at each training epoch. Computing $\mathcal{L}_\text{mc}$ is not also expensive because the number of incorrect classified examples tends to decrease over the training epochs. On the other hand, computing $\mathcal{L}_\text{cc}$ can become very expensive because the number of well-classified instances tends to increase over the training epochs. Therefore, the term $\mathcal{L}_\text{cc}$ should be used wisely, preferably only on challenging problems where just the other two terms of the proposed loss function may not lead to a low training error. 
}
\vspace{-3mm}
\subsection{FCN Architecture} 
\label{sec:filters}
The FCN is a sequence of CLs similar to ones used in CNNs, used as filter banks to learn representation related to textures, as presented in Table~\ref{tab:filters}. Pooling layers follow these layers to reduce the input size progressively. The deeper the layers, the narrower is the latent representation, but with an increasing number of channels, which allows more filters combination, increasing the complexity of the representation.
At the end of the CLs, a GAP layer builds a latent representation where each element represents the response of a filter combination to a texture, measuring how much it happened and not its position on the image. The GAP layer also makes the output dimension independent of the input size, and the latent representation will always have the same number of channels at the end of the FCN. Such a  latent representation feeds a large-margin discriminant to learn classification tasks. We compared our approach with two CNNs with identical architecture (Table~\ref{tab:filters}), but with a sequence of fully connected (FC) layers as discriminant after the GAP layer. We employed binary cross-entropy (BCE) loss and Hinge loss for binary problems and cross-entropy loss in multiclass. 

\begin{table}[h]
\centering
\scriptsize
\setlength{\tabcolsep}{0.6em} 
{\renewcommand{\arraystretch}{1.0}
\caption{FCN used on the LMFCN and in the CNN comparison. ($\textit{w}$: width, $\textit{h}$: height, $\textit{c}$: number of channels, $\phi$: dimension of the latent representation.)
}
\vspace*{-3mm}
\begin{tabular}{rrr}
\hline
Layer & Input & Output \\
\hline
Convolutional Layer & $\textit{w} \times \textit{h} \times \textit{c}$ & $\textit{w} \times \textit{h} \times 64$ \\
Max Pooling & $\textit{w} \times \textit{h} \times 64$ & $\textit{w}/2 \times \textit{h}/2 \times 64$ \\
Convolutional Layer & $\textit{w}/2 \times \textit{h}/2 \times 64$ & $\textit{w}/2 \times \textit{h}/2 \times 128$ \\
Max Pooling & $\textit{w}/2 \times \textit{h}/2 \times 128$ & $\textit{w}/4 \times \textit{h}/4 \times 128$ \\
Convolutional Layer & $\textit{w}/4 \times \textit{h}/4 \times 128$ & $\textit{w}/4 \times \textit{h}/4 \times \phi$ \\
Global Average Pooling & $\textit{w}/4 \times \textit{h}/4 \times \phi$ & $1 \times 1 \times \phi$ \\
\hline
\multicolumn{3}{l}{Batch Normalization and ReLU after each CL.}
\end{tabular}
\label{tab:filters}
}
\vspace*{-4mm}
\end{table}

\label{sec:classifier}

The large margin classifier of the LMFCN is a support vector machine (SVM) with an RBF kernel, which Gram matrix $\mathbf{K}$ is obtained by Eq.~\eqref{eq:kernel} calculated jointly with the distance matrix $\mathbf{D}$. Part of the classifier calculation is reused and performed in GPU. We chose the precomputed RBF kernel due to its ease of computation using the GPU resources and space separation capacity. Although a linear kernel has reduced computational cost, it would require more training of the FCN weights to provide features with more class separation. In the preliminary studies, we compared the two kernel approaches and verified the advantage of the RBF. 

A large-margin discriminant is inherently binary, and to deal with multiclass problems, we adopted the one-vs-all (OVA) approach, which reduces multiclass problems into multiple binary classification problems. In the training stage, one provides all instances to all $\mathit{n_\text{c}}$ pairs of LMFCNs.
After training all the $\mathit{n_\text{c}}$ LMFCNs, we discard the discriminants, keeping only the FCNs. Then, we train a new multiclass SVM. The new classifier is trained using a latent representation with $\phi \times\mathit{n_\text{c}}$, which is the concatenation of all FCN latent representations. At the end of the training process, the full model comprises $\mathit{n_\text{c}}$ FCNs with an $\phi$-dimensional output and a multiclass SVM with a latent representation of $\phi\times\mathit{n_\text{c}}$ dimensions. The multiclass SVM holds internally multiple binary SVMs with RBF kernel on OVA configuration. Although this approach seems similar to using multiple SVMs trained with the FCNs, the new classifiers have access to a latent representation that is better fitted for them. 

The computational cost for training the LMFCN and equivalent CNNs up to the GAP layer is proportional to the number and resolution of input images ($\mathit{n}\times\mathit{w}\times\mathit{h}\times\mathit{c}$) and the number weights ($\alpha_\text{fcn}$). The LMFCN replaces the FC layers (computational cost of $\alpha_\text{fcn}\times\mathit{n}$), by an 
SVM, which requires kernel calculation ($\mathit{n^2}$), and sequential minimal optimization (SMO), which requires $\mathit{n^3}$ in the worst case.
Assuming a small training set, the SMO cost
is lower than $\mathit{n}\times\alpha_\text{fcn}$. However, for large datasets, where $\mathit{n^2}$ is greater than $\alpha_\text{fcn}$, the problem becomes more suitable for conventional CNN architectures. 

The main advantage of the LMFCN is
using only the SVs in the backpropagation, which reduces its computational cost from $\mathit{n}$ to $|\mathbfit{S}|$. The CLs used on both LMFCN and conventional CNNs have several trainable parameters. Therefore, having only SVs as training instances reduces the number of training instances and speeds up the training, making the loss function converge faster within a few epochs. 
\vspace{-4mm}

\section{Experimental Results and Discussion}
\label{sec:results}
\vspace{-3mm}

The evaluation of the proposed LMFCN is carried out on images with texture characteristics and low training data availability. We used a synthetic dataset of images generated from two Gaussian distributions with striped patterns, misc and fabric categories of the Salzburg Texture Image Database~\cite{salzburg},  BreaKHis~\cite{spanhol2016}, BACH~\cite{bach2018} and CRC~\cite{crc} datasets, which comprise histopathological images (HIs). 

\begin{figure*}[ht!]
    \centering
    \includegraphics[width=0.85\linewidth]{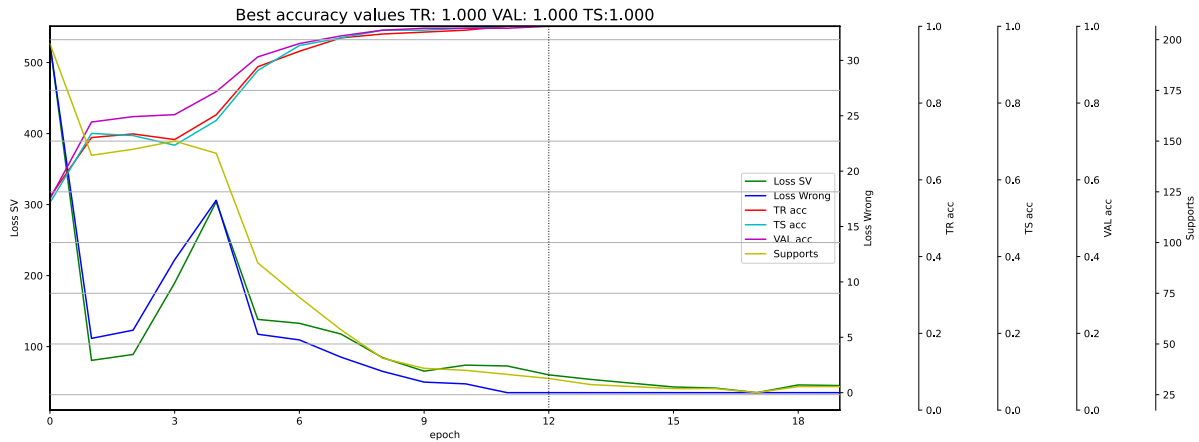}
    \vspace*{-2mm}
    \caption{Training of the LMFCN using synthetic Gaussian images with {a 2-dimensional latent representation. Parameters: $\mathit{sv_\text{close}}$=5, $\mathit{wr_\text{close}}$=1, and $\mathit{sh_\text{close}}$=0.}}
    \vspace*{-5mm}
    \label{fig:gaussian_images_acc}
\end{figure*}

Fig.~\ref{fig:gaussian_images_acc} shows a training graph of the Gaussian images dataset over 20 epochs with the LMFCN. The best-balanced accuracy values for train, validation, and test correspond to the peak validation accuracy. It is possible to observe that the value of losses and the number of SVs goes down over the epochs, and the accuracy rises. That shows that the SVs are getting closer to their instances due to the loss reduction. In addition, the decrease in the number of SVs indicates that a better separability is being achieved. 

\begin{table}[ht!]
\centering
\scriptsize
\setlength{\tabcolsep}{0.35em}
{\renewcommand{\arraystretch}{1.2}
\caption{Balanced accuracy for the LMFCN and equivalent CNN architectures. Epoch refers to training epoch where the best validation accuracy was achieved. NA: Not Applicable, for models that are trained in one single epoch}
\label{tab:results_fcnsv_cnns}
\vspace*{-3mm}
\begin{tabular}{rlcccc}
\hline
Dataset  & Architecture & Training & Validation & Test & Epoch \\
 \hline
\parbox[t]{2mm}{\multirow{5}{*}{\rotatebox[origin=c]{90}{BACH}}} & LMFCN & 0.8811 $\pm$ 0.0426 & 0.7857 $\pm$ 0.0221 & \underline{0.8056 $\pm$ 0.0203} & 6 \\
 & CNN-CE & 0.8926 $\pm$ 0.0135 & \underline{0.8200 $\pm$ 0.0291} & 0.7864 $\pm$ 0.0212 & 88 \\
 & CNN-H & 0.8534 $\pm$ 0.0410 & 0.8038 $\pm$ 0.0261 & 0.7838 $\pm$ 0.0242 & 68 \\
 & ResNet18 & \underline{0.9981 $\pm$ 0.0026} & 0.7596 $\pm$ 0.0122 & 0.7584 $\pm$ 0.0270 & {\color{black}NA}\\
 & InceptionV3 & 0.6295 $\pm$ 0.0746 & 0.5924 $\pm$ 0.0625 & 0.6049 $\pm$ 0.0630 & {\color{black}NA} \\
 \hline
\parbox[t]{2mm}{\multirow{5}{*}{\rotatebox[origin=c]{90}{BreaKHis}}} & LMFCN & \underline{0.9882 $\pm$ 0.0064} & \underline{0.9442 $\pm$ 0.0137} & \underline{0.8942 $\pm$ 0.0372} & 5 \\
 & CNN-CE & 0.8926 $\pm$ 0.0092 & 0.9122 $\pm$ 0.0226 & 0.8926 $\pm$ 0.0140 & 91 \\
 & CNN-H & 0.8560 $\pm$ 0.0041 & 0.8856 $\pm$ 0.0166 & 0.8638 $\pm$ 0.0042 & 91 \\
 & ResNet18 & 0.9777 $\pm$ 0.0062 & 0.8034 $\pm$ 0.0120 & 0.8262 $\pm$ 0.0122 & {\color{black}NA}\\
 & InceptionV3 & 0.6058 $\pm$ 0.0207 & 0.6064 $\pm$ 0.0209 & 0.5949 $\pm$ 0.0207 & {\color{black}NA}\\
 \hline
\parbox[t]{2mm}{\multirow{5}{*}{\rotatebox[origin=c]{90}{CRC}}} & LMFCN & 0.9924 $\pm$ 0.0046 & 0.9914 $\pm$ 0.0024 & 0.9914 $\pm$ 0.0060 & 6 \\
 & CNN-CE & 0.9828 $\pm$ 0.0070 & \underline{0.9934 $\pm$ 0.0025} & 0.9880 $\pm$ 0.0111 & 31 \\
 & CNN-H & 0.9928 $\pm$ 0.0038 & 0.9924 $\pm$ 0.0027 & \underline{0.9928 $\pm$ 0.0070} & 30 \\
 & ResNet18 & \underline{0.9991 $\pm$ 0.0013} & 0.9683 $\pm$ 0.0081 & 0.9680 $\pm$ 0.0150 & {\color{black}NA}\\
 & InceptionV3 & 0.7718 $\pm$ 0.0352 & 0.7670 $\pm$ 0.0375 & 0.7842 $\pm$ 0.0355 & {\color{black}NA}\\
 \hline
\parbox[t]{2.5mm}{\multirow{5}{*}{\rotatebox[origin=c]{90}{Salzburg}}} & LMFCN & 0.9842 $\pm$ 0.0049 & \underline{0.9446 $\pm$ 0.0083} & \underline{0.9230 $\pm$ 0.0081} & 8 \\
 & CNN-CE & 0.8966 $\pm$ 0.0103 & 0.8762 $\pm$ 0.0100 & 0.8602 $\pm$ 0.0115 & 91 \\
 & CNN-H & 0.8258 $\pm$ 0.0257 & 0.8292 $\pm$ 0.0277 & 0.8046 $\pm$ 0.0314 & 84 \\
 & ResNet18 & \underline{0.9946 $\pm$ 0.0033} & 0.9046 $\pm$ 0.0221 & 0.8932 $\pm$ 0.0088 & {\color{black}NA}\\
 & InceptionV3 & 0.6680 $\pm$ 0.0332 & 0.6607 $\pm$ 0.0323 & 0.6644 $\pm$ 0.0473 & {\color{black}NA}\\
 \hline
\parbox[t]{2mm}{\multirow{5}{*}{\rotatebox[origin=c]{90}{Gaussian}}} & LMFCN & \underline{1.0000 $\pm$ 0.0000} & \underline{1.0000 $\pm$ 0.0000} & 0.9990 $\pm$ 0.0022 & 2 \\
 & CNN-CE & 0.9950 $\pm$ 0.0061 & \underline{1.0000 $\pm$ 0.0000} & \underline{1.0000 $\pm$ 0.0000} & 72 \\
 & CNN-H & \underline{1.0000 $\pm$ 0.0000} & \underline{1.0000 $\pm$ 0.0000} & \underline{1.0000 $\pm$ 0.0000} & 56 \\
 & ResNet18 & 0.9740 $\pm$ 0.0109 & 0.6126 $\pm$ 0.0198 & 0.6176 $\pm$ 0.0257 & {\color{black}NA}\\
 & InceptionV3 & 0.5102 $\pm$ 0.0235 & 0.5311 $\pm$ 0.0185 & 0.5021 $\pm$ 0.0258 & {\color{black}NA}\\
 \hline
\end{tabular}
}
\vspace{-7mm}
\end{table}

Table~\ref{tab:results_fcnsv_cnns} compares the results achieved by the LMFCN with other CNN architectures on five datasets. Two CNNs have an architecture similar to the LMFCN, but they use the Hinge loss (CNN-H) and the binary cross-entropy loss (CNN-CE). {These two CNNs are equivalent to TCNNs~\citep{andrearczyk2016} since they use a short sequence of CLs and a GAP to produce energy values.} We also used pretrained ResNet18 and InceptionV3 as feature extractors. We limited the number of training epochs to 100 and 15 epochs for the CNNs and LMFCN. The discriminant layers of the ResNet18 and the InceptionV3 have the same parameters as the LMFCN. Furthermore, we used PCA to reduce the feature vectors generated by these two CNNs from 512 and 2048 to 16, the same size used by the other architectures. Overall, the LMFCN converges to a latent representation that generalizes well much faster (less than ten epochs for all datasets) than different CNN architectures. Furthermore, in 3 out of 5 datasets, the accuracy achieved by the LMFCN on the test sets is higher than that achieved by other CNNs. For the two other datasets, the difference in accuracy is almost negligible. 

Table~\ref{tab:results_acc_bal_trad} shows average results of balanced accuracy of five repetitions comparing the results achieved by the LMFCN with shallow approaches that employ an SVM and three handcrafted feature extractors: Local Binary Pattern (LBP), Gray Level Co-occurrence Matrix (GLCM), and Parameter Free Threshold Adjacency Statistics (PFTAS)~\cite{spanhol2016}. 

For a fair comparison, the LMFCN uses a 59-dimensional latent representation, the same dimension of the smallest feature vector, obtained with the LBP with uniform patterns. PFTAS and GLCM produce 162- and 169-dimensional feature vectors, respectively. PFTAS achieved the best accuracy among the handcrafted feature extractors on three datasets (BACH, BreaKHis, and CRC). LBP achieved the best accuracy on Salzburg and GLCM on the Gaussian dataset. However, the LMFCN with a 59-dimensional latent representation achieves an accuracy higher than all shallow methods on all datasets, indicating its adaptability to different problems and datasets.%

\begin{table}[ht!]
\centering
\scriptsize
\setlength{\tabcolsep}{0.5em}
{\renewcommand{\arraystretch}{1.0}
\caption{Balanced accuracy for LMFCN, GLCM, LBP and PFTAS. The best results on each subset are underlined.}
\vspace*{-4mm}
\begin{tabular}{rlccc}
\hline
Dataset & Method & Training & Validation & Test \\
\hline
\parbox[t]{2mm}{\multirow{5}{*}{\rotatebox[origin=c]{90}{BACH}}} & LMFCN-59 & 0.9664 $\pm$ 0.0751 & 0.7888 $\pm$ 0.0577 & 0.7684 $\pm$ 0.0435 \\
& LMFCN-162 & \underline{1.0000 $\pm$ 0.000} & \underline{0.8068 $\pm$ 0.0600} & \underline{0.7934 $\pm$ 0.0536} \\
& GLCM & 0.7346 $\pm$ 0.0199 & 0.6309 $\pm$ 0.0346 & 0.6618 $\pm$ 0.0711 \\
& LBP & 0.9146 $\pm$ 0.0154 & 0.7455 $\pm$ 0.0442 & 0.7005 $\pm$ 0.0228 \\
& PFTAS & 0.9899 $\pm$ 0.0037 & 0.7372 $\pm$ 0.0634 & 0.7608 $\pm$ 0.0361 \\
\hline
\parbox[t]{2mm}{\multirow{5}{*}{\rotatebox[origin=c]{90}{BreaKHis}}} & LMFCN-59 & \underline{1.0000 $\pm$ 0.0000} & 0.9639 $\pm$ 0.0247 & 0.9476 $\pm$ 0.0075 \\
& LMFCN-162 & \underline{1.0000 $\pm$ 0.0000} & \underline{0.9666 $\pm$ 0.0183} & \underline{0.9595 $\pm$ 0.0050} \\
& GLCM & 0.8128 $\pm$ 0.0086 & 0.8117 $\pm$ 0.0375 & 0.8076 $\pm$ 0.0071 \\
& LBP & 0.9292 $\pm$ 0.0059 & 0.7998 $\pm$ 0.0099 & 0.7925 $\pm$ 0.0098 \\
& PFTAS & 0.9801 $\pm$ 0.0034 & 0.9237 $\pm$ 0.0236 & 0.9145 $\pm$ 0.0168 \\
\hline
\parbox[t]{2mm}{\multirow{4}{*}{\rotatebox[origin=c]{90}{CRC}}} & LMFCN-59 & 0.9977 $\pm$ 0.0022 & \underline{0.9937 $\pm$ 0.0023} & \underline{0.9883 $\pm$ 0.0084} \\
& GLCM & 0.9864  $\pm$ 0.0036 & 0.9828 $\pm$ 0.0055 & 0.9853 $\pm$ 0.0101 \\
& LBP & 0.9974 $\pm$ 0.0012 & 0.9431 $\pm$ 0.0099 & 0.9479 $\pm$ 0.0061 \\
& PFTAS & \underline{1.0000 $\pm$ 0.0000} & 0.9852 $\pm$ 0.0025 & 0.9872 $\pm$ 0.0097 \\
\hline
\parbox[t]{2mm}{\multirow{4}{*}{\rotatebox[origin=c]{90}{Salzburg}}} & LMFCN-59 & \underline{0.9994 $\pm$ 0.0006} & \underline{0.9672 $\pm$ 0.0112} & \underline{0.9500 $\pm$ 0.0180} \\
& GLCM & 0.7706 $\pm$ 0.0163 & 0.7257 $\pm$ 0.0186 & 0.7338 $\pm$ 0.0142 \\
& LBP & 0.9987 $\pm$ 0.0012 & 0.9282 $\pm$ 0.0209 & 0.9116 $\pm$ 0.0208 \\
& PFTAS & 0.9650 $\pm$ 0.0051 & 0.9096 $\pm$ 0.0124 & 0.8973 $\pm$ 0.0137 \\
\hline
\parbox[t]{2mm}{\multirow{4}{*}{\rotatebox[origin=c]{90}{Gaussian}}} & LMFCN-59 & \underline{1.0000 $\pm$ 0.0000} & \underline{0.9354 $\pm$ 0.1254} & \underline{0.9880 $\pm$ 0.0121} \\
& GLCM & 0.9583 $\pm$ 0.0028 & 0.9327 $\pm$ 0.0110 & 0.9527 $\pm$ 0.0191 \\
& LBP & 0.6303 $\pm$ 0.0091 & 0.6710 $\pm$ 0.0131 & 0.6109 $\pm$ 0.0075 \\
& PFTAS & 0.5578 $\pm$ 0.0080 & 0.5953 $\pm$ 0.0077 & 0.5608 $\pm$ 0.0076 \\
\hline
\end{tabular}
\label{tab:results_acc_bal_trad}}
\vspace{-5mm}
\end{table}

The multiclass experiments were carried out on three HI datasets. Our experiments also included comparisons against handcrafted feature extractors GLCM, LBP, and PFTAS as though as ResNet18 and InceptionV3 as feature extractors. We also used a CNN with cross-entropy loss (CE) and similar architecture to the LMFCN but using more filters. The number of filters is proportional to the total number of parameters summing all OVA models of the LMFCN, providing a fair comparison. 

Table~\ref{tab:multi_bach} shows the average balanced accuracy for all evaluated methods. It is noticeable the superior performance of the LMFCN over all others. In the case of the CNN experiments, we allowed the training procedure to extend over 400 epochs. Although the LMFCN uses multiple models, we let each one only train for ten epochs. The metric used, balanced accuracy, helps identify problems on imbalanced datasets. Our approach presented a promising performance on BreaKHis, which has a significant difference between some classes, e.g., Ductal Carcinoma and Phyllodes Tumor. This result shows that the LMFCN performed well in imbalanced scenarios. We also noticed that the LMFCN performed well on the imbalanced OVA subproblems. 

\begin{table}[ht!]
\centering
\scriptsize
\setlength{\tabcolsep}{0.5em}
{\renewcommand{\arraystretch}{1.0}
\caption{Average balanced accuracy and standard deviation for seven methods considering a multiclass scenario.}
\vspace*{-4mm}
\begin{tabular}{clccc}
\hline
 Dataset & Method & \multicolumn{1}{c}{Training} & \multicolumn{1}{c}{Validation} & \multicolumn{1}{c}{Test} \\
 \hline
\parbox[t]{2mm}{\multirow{7}{*}{\rotatebox[origin=c]{90}{BACH}}} & LMFCN & \underline{0.9690 $\pm$ 0.0348} & \underline{0.7170 $\pm$ 0.0210} & \underline{0.6854 $\pm$ 0.0278} \\
& CNN & 0.8390 $\pm$ 0.0184 & 0.6788 $\pm$ 0.0406 & 0.6444 $\pm$ 0.0187 \\
& GLCM & 0.4081 $\pm$ 0.0183 & 0.4003 $\pm$ 0.0753 & 0.3637 $\pm$ 0.0303 \\
& LBP & 0.4643 $\pm$ 0.0334 & 0.4871 $\pm$ 0.0729 & 0.3951 $\pm$ 0.0896 \\
& PFTAS & 0.6229 $\pm$ 0.0146 & 0.5690 $\pm$ 0.0464 & 0.6043 $\pm$ 0.0235 \\
& ResNet18 & 1.0000 $\pm$ 0.0000 & 0.6056 $\pm$ 0.0578 & 0.5671 $\pm$ 0.0300 \\
& InceptionV3 & 0.4628 $\pm$ 0.0807 & 0.4269 $\pm$ 0.0904 & 0.4057 $\pm$ 0.0261 \\
\hline
\parbox[t]{2mm}{\multirow{7}{*}{\rotatebox[origin=c]{90}{BreaKHis}}} & LMFCN & \underline{0.9688 $\pm$ 0.0285} & \underline{0.8125 $\pm$ 0.0321} & \underline{0.7895 $\pm$ 0.0162} \\
& CNN & 0.8032 $\pm$ 0.0319 & 0.7347 $\pm$ 0.0238 & 0.7040 $\pm$ 0.0307 \\
& GLCM & 0.4697 $\pm$ 0.0111 & 0.4218 $\pm$ 0.0212 & 0.4091 $\pm$ 0.0050 \\
& LBP & 0.9235 $\pm$ 0.0084 & 0.5373 $\pm$ 0.0174 & 0.5218 $\pm$ 0.0147 \\
& PFTAS & 0.9537 $\pm$ 0.0086 & 0.6643 $\pm$ 0.0109 & 0.6768 $\pm$ 0.0147 \\
& ResNet18 & 1.0000 $\pm$ 0.0000 & 0.5933 $\pm$ 0.0341 & 0.5834 $\pm$ 0.0213 \\
& InceptionV3 & 0.1848 $\pm$ 0.0182 & 0.1745 $\pm$ 0.0194 & 0.1767 $\pm$ 0.0123 \\
\hline
\parbox[t]{2mm}{\multirow{7}{*}{\rotatebox[origin=c]{90}{CRC}}} & LMFCN & \underline{0.9834 $\pm$ 0.0127} & \underline{0.9379 $\pm$ 0.0065} & \underline{0.9338 $\pm$ 0.0073} \\
& CNN & 0.7209 $\pm$ 0.2614 & 0.3946 $\pm$ 0.0453 & 0.3901 $\pm$ 0.0402 \\
& GLCM & 0.6062 $\pm$ 0.0057 & 0.5960 $\pm$ 0.0069 & 0.6049 $\pm$ 0.0084 \\
& LBP & 0.6148 $\pm$ 0.0045 & 0.6086 $\pm$ 0.0270 & 0.6144 $\pm$ 0.0123 \\
& PFTAS & 0.8359 $\pm$ 0.0058 & 0.8271 $\pm$ 0.0152 & 0.8297 $\pm$ 0.0097 \\
& ResNet18 & 0.9506 $\pm$ 0.0033 & 0.5070 $\pm$ 0.0141 & 0.5066 $\pm$ 0.0138 \\
& InceptionV3 & 0.1509 $\pm$ 0.0036 & 0.1546 $\pm$ 0.0075 & 0.1477 $\pm$ 0.0082 \\
\hline
\end{tabular}
\label{tab:multi_bach}}
\vspace{-5mm}
\end{table}

The experimental results have shown that the LMFCN outperforms all other methods in a scenario composed of textural images and small-size datasets. Besides achieving higher accuracy, the proposed method has several advantages over the shallow and deep related methods. The LMFCN approach requires few data to train a sequence of CLs and a large-margin discriminant properly. In the experiments comparing the LMFCN with equivalent CNNs, using a 16-dimensional latent representation, the LMFCN achieved training stability and high accuracy within 20 training epochs. The CNNs needed 100 epochs to achieve comparable performance. 

The latent representation learned by the LMFCN, constrained to a dimensionality similar to shallow methods (59- and 162-dimensional), is more discriminant than LBP, PFTAS, and GLCM. As a result, the LMFCN achieved higher balanced accuracy than the compared methods. Furthermore, the LMFCN obtained a competitive accuracy even with a 16-dimensional latent representation. Moreover, the improvement achieved by increasing the latent representation to 59 dimensions is meaningful, with a slight gain when increasing it up to 169. 

The LMFCN reduces the number of SVs over the training epochs. Fewer SVs imply that the computational effort reduces, as the main term of the loss function depends on the number of SVs. Compared to equivalent CNNs, the LMFCN has an extra cost related to the computation of kernel $\mathbf{K}$, distance matrice $\mathbf{D}$, and the quadratic optimization problem solved by the SMO algorithm, which have computational complexities $\mathcal{O}(n^2)$, $\mathcal{O}(n^2)$, and $\mathcal{O}(n^3)$, respectively. CNNs have a training complexity proportional to the number of instances and weights. They have more parameters, use more instances in the backpropagation, and take more epochs to converge than the LMFCN. 

Data imbalance is burdensome for training machine learning algorithms. However, the LMFCN deals well with imbalanced classes in two-class and multi-class scenarios, achieving competitive performance even on highly imbalanced OVA subproblems. Furthermore, the computational effort is not extremely high, given the reduced number of epochs to train the model at each subproblem.
\vspace{-4mm}
\section{Conclusion}
\label{sec:conclusion}
\vspace{-3mm}
This paper proposed a large-margin representation learning approach that is made up of convolutional layers and a large-margin discriminant. As a result, the LMFCN achieved competitive accuracy compared to CNNs with similar architecture while reducing the computational cost. Furthermore, the LMFCN achieved training stability in a few epochs thanks to using only the SVs in the backpropagation algorithm. These achievements were possible using a large-margin discriminant, which replaces the fully connected and softmax layers of conventional CNNs. As a result, an SVM with an RBF kernel can produce complex margins, avoiding expensive refinement of the latent representation. This way, the FCNs do not need to suffer drastic updates.
Unlike the handcrafted features extractors, the LMFCN has more adaptability given its consistent results in different datasets. The most problematic scenario for the LMFCN is the multiclass classification task, but it showed promising results despite the overhead caused by the OVA approach. We used a few epochs per subproblem to alleviate the computational cost of using several models, keeping the cost of our approach similar to the CNNs.
In conclusion, the LMFCN has advantages in computational cost and classification performance over the compared methods for small datasets of textural images.

\vspace{-4mm}
\section*{Acknowledgments}
\vspace{-3mm}
This work was funded by the Natural Sciences and Engineering Research Council of Canada (NSERC) Grant RGPIN 2016-04855.

\vspace{-4mm}
\bibliography{mybibfile}

\begin{thebibliography}{15}
\expandafter\ifx\csname natexlab\endcsname\relax\def\natexlab#1{#1}\fi
\providecommand{\url}[1]{\texttt{#1}}
\providecommand{\href}[2]{#2}
\providecommand{\path}[1]{#1}
\providecommand{\DOIprefix}{doi:}
\providecommand{\ArXivprefix}{arXiv:}
\providecommand{\URLprefix}{URL: }
\providecommand{\Pubmedprefix}{pmid:}
\providecommand{\doi}[1]{\href{http://dx.doi.org/#1}{\path{#1}}}
\providecommand{\Pubmed}[1]{\href{pmid:#1}{\path{#1}}}
\providecommand{\bibinfo}[2]{#2}
\ifx\xfnm\relax \def\xfnm[#1]{\unskip,\space#1}\fi
\bibitem[{He et~al.(2016)He, Zhang, Ren, and Sun}]{resnet}
\bibinfo{author}{K.~He}, \bibinfo{author}{X.~Zhang}, \bibinfo{author}{S.~Ren},
  \bibinfo{author}{J.~Sun},
\newblock \bibinfo{title}{Deep residual learning for image recognition},
\newblock in: \bibinfo{booktitle}{IEEE/CVF Conf Comp Vis Patt Recog},
  \bibinfo{year}{2016}, pp. \bibinfo{pages}{770--778}.
\bibitem[{Szegedy et~al.(2016)Szegedy, Vanhoucke, Ioffe, Shlens, and
  Wojna}]{inception}
\bibinfo{author}{C.~Szegedy}, \bibinfo{author}{V.~Vanhoucke},
  \bibinfo{author}{S.~Ioffe}, \bibinfo{author}{J.~Shlens},
  \bibinfo{author}{Z.~Wojna},
\newblock \bibinfo{title}{Rethinking the inception architecture for computer
  vision},
\newblock in: \bibinfo{booktitle}{IEEE/CVF Conf Comp Vis Patt Recog},
  \bibinfo{year}{2016}, pp. \bibinfo{pages}{2818--2826}.
\bibitem[{Russakovsky et~al.(2015)Russakovsky, Deng, Su, Krause, Satheesh, Ma,
  Huang, Karpathy, Khosla, Bernstein, Berg, and Fei-Fei}]{imagenet}
\bibinfo{author}{O.~Russakovsky}, \bibinfo{author}{J.~Deng},
  \bibinfo{author}{H.~Su}, \bibinfo{author}{J.~Krause},
  \bibinfo{author}{S.~Satheesh}, \bibinfo{author}{S.~Ma},
  \bibinfo{author}{Z.~Huang}, \bibinfo{author}{A.~Karpathy},
  \bibinfo{author}{A.~Khosla}, \bibinfo{author}{M.~Bernstein},
  \bibinfo{author}{A.~C. Berg}, \bibinfo{author}{L.~Fei-Fei},
\newblock \bibinfo{title}{{ImageNet Large Scale Visual Recognition Challenge}},
\newblock \bibinfo{journal}{Intl Journal of Computer Vision}
  \bibinfo{volume}{115} (\bibinfo{year}{2015}) \bibinfo{pages}{211--252}.
  \DOIprefix\doi{10.1007/s11263-015-0816-y}.
\bibitem[{Andrearczyk and Whelan(2016)}]{andrearczyk2016}
\bibinfo{author}{V.~Andrearczyk}, \bibinfo{author}{P.~F. Whelan},
\newblock \bibinfo{title}{Using filter banks in convolutional neural networks
  for texture classification},
\newblock \bibinfo{journal}{Pattern Recognition Letters} \bibinfo{volume}{84}
  (\bibinfo{year}{2016}) \bibinfo{pages}{63 -- 69}.
  \DOIprefix\doi{https://doi.org/10.1016/j.patrec.2016.08.016}.
\bibitem[{{Cimpoi} et~al.(2015){Cimpoi}, {Maji}, and {Vedaldi}}]{cimpoi2015}
\bibinfo{author}{M.~{Cimpoi}}, \bibinfo{author}{S.~{Maji}},
  \bibinfo{author}{A.~{Vedaldi}},
\newblock \bibinfo{title}{Deep filter banks for texture recognition and
  segmentation},
\newblock in: \bibinfo{booktitle}{IEEE Conf Comp Vis Patt Recog},
  \bibinfo{year}{2015}, pp. \bibinfo{pages}{3828--3836}.
\bibitem[{{Humeau-Heurtier}(2019)}]{heurtier2019}
\bibinfo{author}{A.~{Humeau-Heurtier}},
\newblock \bibinfo{title}{Texture feature extraction methods: A survey},
\newblock \bibinfo{journal}{IEEE Access} \bibinfo{volume}{7}
  (\bibinfo{year}{2019}) \bibinfo{pages}{8975--9000}.
\bibitem[{{Chopra} et~al.(2005){Chopra}, {Hadsell}, and {LeCun}}]{chopra2005}
\bibinfo{author}{S.~{Chopra}}, \bibinfo{author}{R.~{Hadsell}},
  \bibinfo{author}{Y.~{LeCun}},
\newblock \bibinfo{title}{Learning a similarity metric discriminatively, with
  application to face verification},
\newblock in: \bibinfo{booktitle}{IEEE CS Conf Comp Vis Patt Recog},
  volume~\bibinfo{volume}{1}, \bibinfo{year}{2005}, pp.
  \bibinfo{pages}{539--546 vol. 1}. \DOIprefix\doi{10.1109/CVPR.2005.202}.
\bibitem[{Kim et~al.(2020)Kim, Kim, Cho, and Kwak}]{proxyloss2020}
\bibinfo{author}{S.~Kim}, \bibinfo{author}{D.~Kim}, \bibinfo{author}{M.~Cho},
  \bibinfo{author}{S.~Kwak},
\newblock \bibinfo{title}{Proxy anchor loss for deep metric learning},
\newblock in: \bibinfo{booktitle}{IEEE/CVF Conf Comp Vis Patt Recog},
  \bibinfo{year}{2020}.
\bibitem[{Wang et~al.(2019)Wang, Hua, Kodirov, Hu, Garnier, and
  Robertson}]{wang2019}
\bibinfo{author}{X.~Wang}, \bibinfo{author}{Y.~Hua},
  \bibinfo{author}{E.~Kodirov}, \bibinfo{author}{G.~Hu},
  \bibinfo{author}{R.~Garnier}, \bibinfo{author}{N.~M. Robertson},
\newblock \bibinfo{title}{Ranked list loss for deep metric learning},
\newblock in: \bibinfo{booktitle}{IEEE/CVF Conf Comp Vis Patt Recog (CVPR)},
  \bibinfo{year}{2019}, pp. \bibinfo{pages}{5207--5216}.
\bibitem[{Movshovitz-Attias et~al.(2017)Movshovitz-Attias, Toshev, Leung,
  Ioffe, and Singh}]{movshovitz2017}
\bibinfo{author}{Y.~Movshovitz-Attias}, \bibinfo{author}{A.~Toshev},
  \bibinfo{author}{T.~K. Leung}, \bibinfo{author}{S.~Ioffe},
  \bibinfo{author}{S.~Singh},
\newblock \bibinfo{title}{No fuss distance metric learning using proxies},
\newblock in: \bibinfo{booktitle}{IEEE Intl Conf on Comp Vis},
  \bibinfo{year}{2017}, pp. \bibinfo{pages}{360--368}.
\bibitem[{Aziere and Todorovic(2019)}]{aziere2019}
\bibinfo{author}{N.~Aziere}, \bibinfo{author}{S.~Todorovic},
\newblock \bibinfo{title}{Ensemble deep manifold similarity learning using hard
  proxies},
\newblock in: \bibinfo{booktitle}{IEEE/CVF Conf Comp Vis Patt Recog},
  \bibinfo{year}{2019}, pp. \bibinfo{pages}{7299--7307}.
\bibitem[{Roland~Kwitt(2019)}]{salzburg}
\bibinfo{author}{P.~M. Roland~Kwitt}, \bibinfo{title}{Stex, salzburg texture
  image database (stex)},
  \bibinfo{howpublished}{\url{https://wavelab.at/sources/STex/}},
  \bibinfo{year}{2019}. \bibinfo{note}{Accessed: 2022-03-28}.
\bibitem[{{Spanhol} et~al.(2016){Spanhol}, {Oliveira}, {Petitjean}, and
  {Heutte}}]{spanhol2016}
\bibinfo{author}{F.~A. {Spanhol}}, \bibinfo{author}{L.~S. {Oliveira}},
  \bibinfo{author}{C.~{Petitjean}}, \bibinfo{author}{L.~{Heutte}},
\newblock \bibinfo{title}{Breast cancer histopathological image classification
  using convolutional neural networks},
\newblock in: \bibinfo{booktitle}{Intl Joint Conf Neural Networks},
  \bibinfo{year}{2016}, pp. \bibinfo{pages}{2560--2567}.
\bibitem[{Aresta et~al.(2019)Aresta, Araújo, Kwok, and et. al.}]{bach2018}
\bibinfo{author}{G.~Aresta}, \bibinfo{author}{T.~Araújo},
  \bibinfo{author}{S.~Kwok}, \bibinfo{author}{et. al.},
\newblock \bibinfo{title}{Bach: Grand challenge on breast cancer histology
  images},
\newblock \bibinfo{journal}{Medical Image Analysis} \bibinfo{volume}{56}
  (\bibinfo{year}{2019}) \bibinfo{pages}{122 -- 139}.
  \DOIprefix\doi{https://doi.org/10.1016/j.media.2019.05.010}.
\bibitem[{Kather et~al.(2016)Kather, Weis, Bianconi, Melchers, Schad, Gaiser,
  Marx, and Z{\"o}llner}]{crc}
\bibinfo{author}{J.~N. Kather}, \bibinfo{author}{C.-A. Weis},
  \bibinfo{author}{F.~Bianconi}, \bibinfo{author}{S.~M. Melchers},
  \bibinfo{author}{L.~R. Schad}, \bibinfo{author}{T.~Gaiser},
  \bibinfo{author}{A.~Marx}, \bibinfo{author}{F.~G. Z{\"o}llner},
\newblock \bibinfo{title}{Multi-class texture analysis in colorectal cancer
  histology},
\newblock \bibinfo{journal}{Scientific Reports} \bibinfo{volume}{6}
  (\bibinfo{year}{2016}) \bibinfo{pages}{27988}.
  \DOIprefix\doi{10.1038/srep27988}.

\end{thebibliography}
\end{document}